\documentclass{kais}
\usepackage{wrapfig,psfig,epsf}
\usepackage[dcucite]{harvard}
\usepackage{graphicx}
\usepackage{minitoc}
\usepackage{amsmath}
\usepackage{footnote}
\usepackage{hyperref}

\pubyear{2000}
\pagerange{\pageref{firstpage}--\pageref{lastpage}}
\volume{xxx}

\begin{document}
\label{firstpage}

\title{Analyzing privacy-aware mobility behavior using the evolution of spatio-temporal entropy}

\author[A. Moro et al.]{Arielle Moro$^1$, Beno\^{i}t Garbinato$^1$ and Val\'erie Chavez-Demoulin$^2$\\ $^1$Department of Information Systems, University of Lausanne, Lausanne, Switzerland;\\ $^2$Department of Operations, University of Lausanne, Lausanne, Switzerland}
 
\maketitle

\begin{abstract}
Analyzing mobility behavior of users is extremely useful to create or improve existing services.
Several research works have been done in order to study mobility behavior of users that mainly use users' significant locations.
However, these existing analysis are extremely intrusive because they require the knowledge of the frequently visited places of users, which thus makes it fairly easy to identify them.
Consequently, in this paper, we present a privacy-aware methodology to analyze mobility behavior of users.
We firstly propose a new metric based on the well-known Shannon entropy, called spatio-temporal entropy, to quantify the mobility level of a user during a time window.
Then, we compute a sequence of spatio-temporal entropy from the location history of the user that expresses user's movements as rhythms.
We secondly present how to study the effects of several groups of additional variables on the evolution of the spatio-temporal entropy of a user, such as spatio-temporal, demographic and mean of transportation variables.
For this, we use Generalized Additive Models (GAMs).
The main strength of GAMs is that they are not only interesting to predict a response variable, but also to understand the effects of co-variables on this response variable.
The results firstly show that the spatio-temporal entropy and GAMs are an ideal combination to understand mobility behavior of an individual user or a group of users.
We also evaluate the prediction accuracy of a global GAM compared to individual GAMs and individual AutoRegressive Integrated Moving Average (ARIMA) models.
These last results highlighted that the global GAM gives more accurate predictions of spatio-temporal entropy by checking the Mean Absolute Error (MAE).
In addition, this research work opens various threads, such as the prediction of demographic data of users or the creation of personalized mobility prediction models by using movement rhythm characteristics of a user.
\end{abstract}


\section{Introduction}
\label{sec:intro}

Studying mobility behavior of users is a key element to understand how they usually move and what are the events that can influence their behavior.
We can study mobility trends from a user point of view with her own movements or from a population point of view with multiple users' movements.
This mobility analysis can be useful for several goals: for example, studying the movements of a population through a city to adapt transportation modes and mobility paths.
It can also be very helpful to predict mobility behavior of a user from other users' movements  having common characteristics.
However, studying mobility behavior can be extremely intrusive and lead to a location privacy issue.

In the literature, mobility behavior has been analyzed using mobility patterns from the extraction of significant places visited by a user as described in \cite{Gonzalez2008,Sadilek2012,Zion2017}.
For example, the most common places studied in mobility analysis are home and work places because users usually spend a large amount of time in them.
However, such analysis is highly intrusive because it is highlighting the most significant places of a user, i.e., where she spends most of her time.
Consequently, it should be possible to find the identity of a user from these significant places.
For example, in \cite{Montjoye2013}, De Montjoye et al. demonstrate that only 4 spatio-temporal points are sufficient to uniquely identify 95\% of the users of the dataset they used for their research.
All these findings indicate that it is crucial to find a new methodology to analyse the mobility behavior of a user or multiple users by only analyzing their rhythm, without necessarily studying their significant visited locations.

In order to reach the goal of this mobility behavior analysis that takes care of the location privacy of users, we propose a new methodology based on the extraction of movement rhythm of a user from her location history. 
To do so, we compute a spatio-temporal entropy sequence that describes the mobility of the user during several time windows of the same duration, e.g., one hour.
Then, we analyze the effects of different variables on this rhythm to better understand the way the user moves.
These variables can belong to various categories and are strongly linked to the availability of data of the chosen dataset we use: spatio-temporal variables (e.g., day of the week, hour of the day, max distance travelled) and demographic variables (e.g., gender, age-group, working-profile, sport exercise frequency).
We use a dataset, called \emph{Breadcrumbs}, that has been collected in the Lake Geneva region (Switzerland) at the Lausanne campus in~2018.
We choose Generalized Additive Models (GAMs) as a support to understand a mobility behavior because they are not only interesting to predict a response variable, but also to understand what kind of co-variables can affect this response variable.
In our context, we obviously use the spatio-temporal entropy as the response variable of the GAM because its value reflects the mobility level of a user during a specific time window.
Hence, the sequence of spatio-temporal entropy of a user describes her movement rhythms.

In order to evaluate the prediction accuracy of the GAMs for our research analysis, we train one global GAM with 60\% of the spatio-temporal entropy sequence of all the users of the dataset, and check the prediction obtained for the 40\% remaining sequence of all of them.
We individual AutoRegressive Integrated Moving Average (ARIMA) models for each user with 60\% of their dataset and compare the Mean Absolute Errors (MAEs) and the Root Mean Squared Errors (RMSEs) obtained of each user for these two different models.
We also compare the MAEs and the RMSEs obtained for individual GAMs compared to those resulting of the use of the global GAM.
We observe that there is not a significant difference between individual GAMs and the global GAM in terms of accuracy and that the global GAM is more accurate than the individual ARIMA models if we compare the MAE results.
In addition, the results show that it is possible to observe the effect of variables on the mobility of a user or a group of users in a privacy-aware manner by using the spatio-temporal entropy only as a response variable.\\


\noindent To the best of our knowledge, this research work has two main contributions: 

\begin{enumerate}
	\item Proposing a new methodology to analyze mobility behavior of users in a privacy manner;
	\item Describing a new entropy-based metric to quantify the level of mobility of a user that expresses users' movements as rhythms.
\end{enumerate}

\noindent The roadmap of this paper is the following. 
In Section~\ref{sec:methodology}, we present the overall privacy-aware methodology to analyze mobility behaviors.
Then, in Section~\ref{sec:from_locations_to_spatiotemporal_entropy}, we present how we build the spatio-temporal entropy sequence of a user from her location history.
In Section~\ref{sec:mobility_behavior_analysis_prediction}, we describe the entire analysis of the mobility using GAMs, ARIMA models and the results obtained with these two types of predictive models.
Section~\ref{sec:related_work} details the related work.
And finally, we conclude the paper in Section~\ref{sec:conclusion} and present the future work.

\section{Proposed methodology}
\label{sec:methodology}

As mentioned in the introduction, the goal of this research work is to analyse the mobility behavior of a user or a group of users in a location-privacy manner, i.e., without extracting sensitive places frequently visited by a user, such as her home place, work place and favorite other places.
In order to be more precise, we want to analyze the mobility behavior of the users according to  variables that can affect it.

The first challenge is to convert the location history of a user into a rhythm.
To do so, we create a metric based on the well-known Shannon entropy in order to quantify the level of mobility of the user at a certain time and through space.
This crucial step must be done for the entire location history of the user in order to obtain a sequence of spatio-temporal entropy that represents a time series.
Parallel to this first step, it is also important to collect several variables that we want to explore during the next step, which is the mobility analysis.
These variables can belong to different categories (e.g., spatio-temporal, demographic and mean of transportation variables) will be used to see their influence on the evolution of the spatio-temporal entropy.
For example, these variables can be the day of the week, the maximum speed of the user during the time window, if a user has a job during the week, etc$\ldots$

The second challenge is now to analyze the mobility behavior of a user or a group of users.
Hence, we must evaluate how the user or the users usually move(s).
To do so, we decide to user Generalized Additive Models (GAMs) that are extremely valuable for this work because these models can not only predict a response variable, which is obviously the spatio-temporal entropy, but also measure the effects of co-variables on this response variable.
A GAM can be formally noted as follows in Equation~\ref{equ:standard_gam}.
Let $Y$ be the spatio-temporal entropy variable and $X_1,...,X_m$ a set of co-variables. 
$\beta_0$ is the intercept and $f_j$, $j=1,\ldots,m$ are smoothed functions.
We consider that we have $n$ observations, such that $\{(y_i,x_{1,i},\ldots,x_{m,i})\}_{i=1}^n$
Finally, $\varepsilon_{i}$ is a random error.

\begin{equation}
	y_{i} = \beta_{0} + f_{1}(x_{1_{i}}) + \ldots + f_{m}(x_{m_{i}}) + \varepsilon_{i}
\label{equ:standard_gam}
\end{equation}

\section{From locations to a spatio temporal entropy sequence}
\label{sec:from_locations_to_spatiotemporal_entropy}

This section presents how we can compute the spatio-temporal entropy sequence from a location history of a user.
Firstly, we describe how is captured the location history of the user, and secondly, we detail its transformation into a sequence of spatio-temporal entropy.

\subsection{Location history}
\label{sec:location_history}

In order to collect the location history of a user, we consider that a user owns a mobile device being able to compute its location at any time via the use of an embedded postitioning system.
In addition, this user is moving on the surface of the earth and can use different means of transport.
A raw location of the user captured by the mobile device has the following notation: ~$loc = (\phi, \lambda, t)$ in which~$\phi$ is a latitude, $\lambda$ is a longitude and $t$ is a timestamp representing when the location was caught on the mobile device.
Consequently, the location history of a user can be expressed as the following sequence of $n$ locations:~$L = \langle loc_{1}, loc_{2}, \cdots, loc_{n}\rangle$.

\begin{figure}
\center
\includegraphics[scale=0.35]{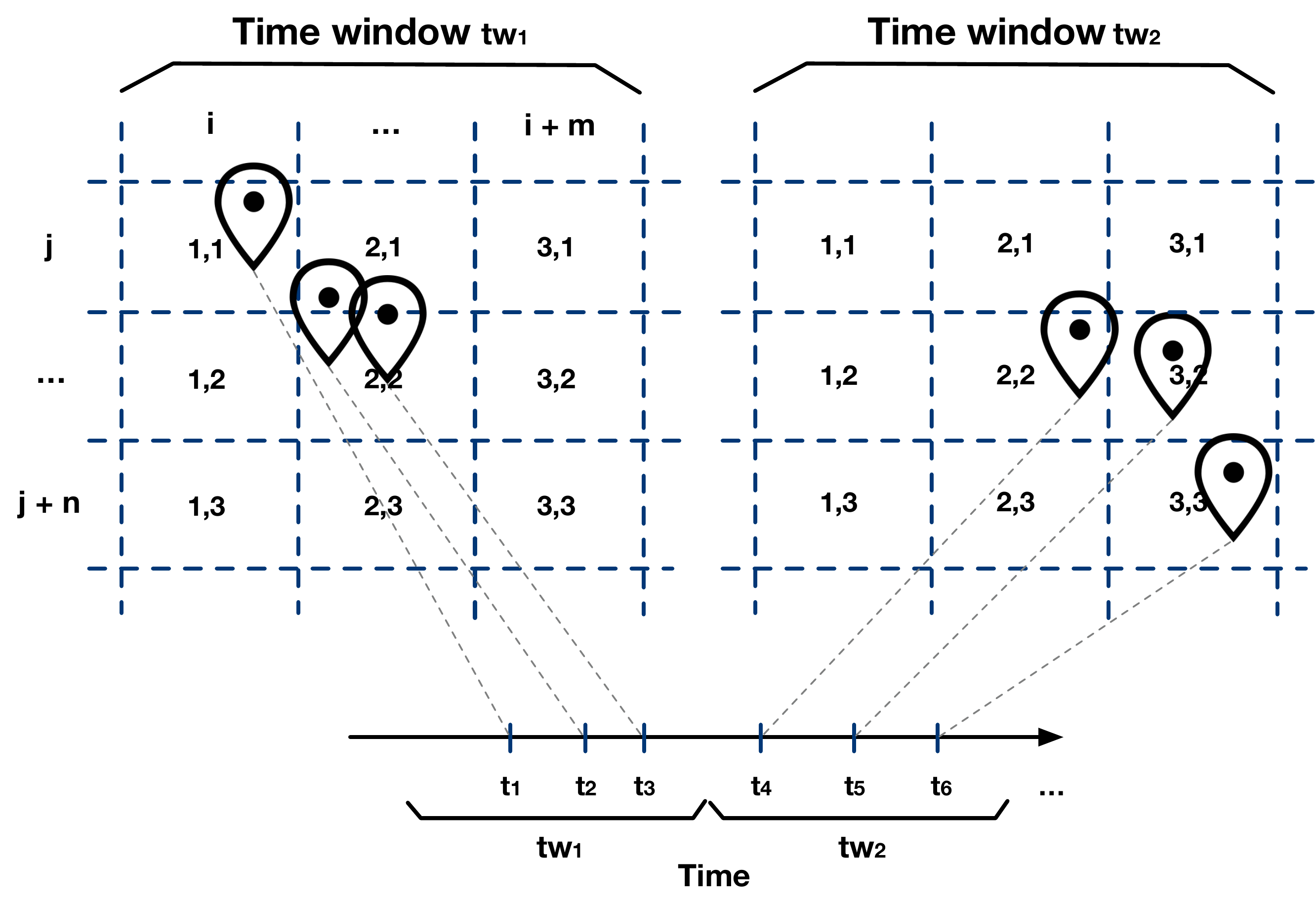}
\caption{Grid structure supporting the computation of the spatio-temporal entropy sequence}
\label{fig:spatiotemporalentropy_computation}
\end{figure}

\subsection{Spatio-temporal entropy sequence}
\label{sec:spatiotemporal_entropy}

To compute the spatio-temporal entropy sequence of a user based on her location history, we firstly need to create a spatial grid to discretize the space in which the user is moving.
The space is divided into a 2-dimensional space in which the position of each cell in this grid is described as a pair $(i,j)$.
We also note $n$ and $m$ the numbers of cells along the x-axis and along the y-axis respectively.
We divide the time duration of the location history of the user into $T$ time windows having the same period of time, e.g., 3600 seconds (one hour).
Figure~\ref{fig:spatiotemporalentropy_computation} depicts the state of the grid for two successive time windows.
Then, we compute the time proportion $p_{i,j}^{tw_{k}}$ for each visited cell of the grid by the user during the specific time window $tw_{k}$, in which $k$ is the $k$th time window computed.

Before describing the entropy computation of a specific time window, we detail the computation of the time proportion below.\\

\noindent\textbf{Time proportion computation.}\\

\begin{figure}
\center
\includegraphics[scale=0.4]{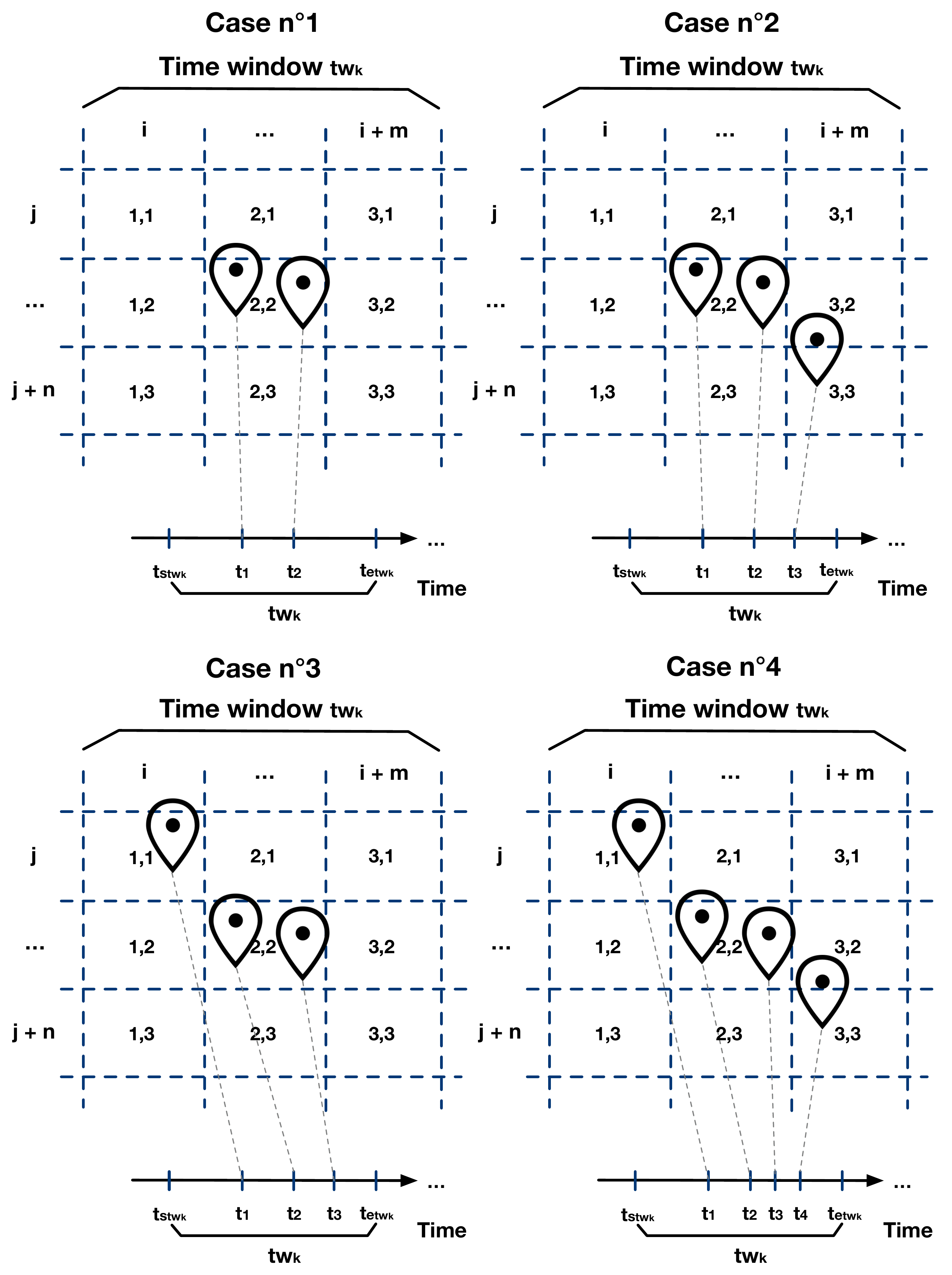}
\caption{Time proportion $p_{i,j}^{tw_{k}}$ computation for a specific time window $tw_{k}$}
\label{fig:spatio_temporal_entropy_time_proportion_computation_cases}
\end{figure}

The computation of a time proportion $p_{i,j}$ spent in a cell $(i,j)$ of the grid visited by the user during a specific time window is linked to all successive locations visited in this cell during this specific time window.
In order to support our explanation, Figure~\ref{fig:spatio_temporal_entropy_time_proportion_computation_cases} presents the context of time proportion computation of two or three visited cells of the grid.

This part aims at explaining the four different cases that we can encounter when we must compute the time proportion of visited cells during a time window $tw_{k}$ by a user.
The first case describes a simple case in which we can consider that the user spent the entire time window $tw_{k}$ in the cell $(2,2)$.
In the second case, we assume that the user was in cell $(2,2)$ since the beginning of the time window $tw_{k}$ and then in the cell $(3,3)$ until the end of $tw_{k}$. 
The time spent between the second location and the third location is equally distributed between $(2,2)$ and $(3,3)$.
The third case is similar to the second case. However, the time spent between the first location and the second location is equally distributed between $(1,1)$ and $(2,2)$.
Finally, the fourth case is the combination of the last two cases.

In order to formally describe the four cases, we introduce the following elements below.
Firstly, we introduce a first new sequence of locations~$L_{tw_{k}}$ that contains the successive locations visited by the user during the time window $tw_{k}$, which is obviously a subsequence of the location history of the user~$L$.
Secondly, we introduce a second new sequence of locations~$L_{i,j_{tw_{k}}}$ containing all successive locations belonging to the same cell that are visited during the time window~$tw_{k}$, which is obviously a subsequence of the location history of the user~$L_{tw_{k}}$.
Finally, if the subsequence of locations~$L_{i,j_{tw_{k}}}$ contains more than one location, there are four specific cases described in Equation~\ref{equ:time_proportion_computation1}, \ref{equ:time_proportion_computation2}, \ref{equ:time_proportion_computation3} and~\ref{equ:time_proportion_computation4} below in order to illustrate the four specific cases described below.
For sake of simplicity, we add $.f$ or $.l$ to the subsequences introduced above in order to identify the first or the last location(s) of them respectively.
We also assume that $L_{i,j_{tw_{k}}}.f$ corresponds to $loc_{b}$ in the sequence $L_{tw_{k}}$ and $L_{i,j_{tw_{k}}}.l$ corresponds to $loc_{c}$ in the sequence $L_{tw_{k}}$.
In addition, we define $t_{s_{tw_{k}}}$ and $t_{e_{tw_{k}}}$ as the starting and ending timestamp of the time window $tw_{k}$ respectively.\\

\noindent\textbf{First case:} if $L_{tw_{k}}.f$ is equal to $L_{i,j_{tw_{k}}}.f$ and if $L_{tw_{k}}.l$ is equal to $L_{i,j_{tw_{k}}}.l$.

\begin{equation}
	p_{i,j}^{tw_{k}} = \frac{(L_{i,j_{tw_{k}}}.f.t - t_{s_{tw_{k}}}) + (L_{i,j_{tw_{k}}}.l.t - L_{i,j_{tw_{k}}}.f.t) + (t_{e_{tw_{k}}} - L_{i,j_{tw_{k}}}.l.t)}{t_{e_{tw_{k}}} - t_{s_{tw_{k}}}}
\label{equ:time_proportion_computation1}
\end{equation}

\noindent\textbf{Second case:} if $L_{tw_{k}}.f$ is equal to $L_{i,j_{tw_{k}}}.f$ and if $L_{tw_{k}}.l$ is not equal to $L_{i,j_{tw_{k}}}.l$.

\begin{equation}
	p_{i,j}^{tw_{k}} = \frac{(L_{i,j_{tw_{k}}}.f.t - t_{s_{tw_{k}}}) + (L_{i,j_{tw_{k}}}.l.t - L_{i,j_{tw_{k}}}.f.t) + ((loc_{c}.t - loc_{c+1}.t)/2.0)}{t_{e_{tw_{k}}} - t_{s_{tw_{k}}}}
\label{equ:time_proportion_computation2}
\end{equation}

\noindent\textbf{Third case:} if $L_{tw_{k}}.f$ is not equal to $L_{i,j_{tw_{k}}}.f$ and if $L_{tw_{k}}.l$ is equal to $L_{i,j_{tw_{k}}}.l$.

\begin{equation}
	p_{i,j}^{tw_{k}} = \frac{((loc_{b-1}.t - loc_{b}.t)/2.0) + (L_{i,j_{tw_{k}}}.l.t - L_{i,j_{tw_{k}}}.f.t) +(t_{e_{tw_{k}}} - L_{i,j_{tw_{k}}}.l.t)}{t_{e_{tw_{k}}} - t_{s_{tw_{k}}}}
\label{equ:time_proportion_computation3}
\end{equation}

\noindent\textbf{Fourth case:} if $L_{tw_{k}}.f$ is not equal to $L_{i,j_{tw_{k}}}.f$ and if $L_{tw_{k}}.l$ is not equal to $L_{i,j_{tw_{k}}}.l$.

\begin{equation}
	p_{i,j}^{tw_{k}} = \frac{((loc_{b-1}.t - loc_{b}.t)/2.0) + (L_{i,j_{tw_{k}}}.l.t - L_{i,j_{tw_{k}}}.f.t) + ((loc_{c}.t - loc_{c+1}.t)/2.0)}{t_{e_{tw_{k}}} - t_{s_{tw_{k}}}}
\label{equ:time_proportion_computation4}
\end{equation}

If there is no location in the cells of the grid visited by the user during the time window, i.e., in~$L_{tw_{k}}$, the entropy of this time window takes the simple value of \emph{"NA"}.\\

\noindent\textbf{Spatio-temporal entropy computation.}\\

Finally, the entropy computation of a specific time window is detailed in Equation~\ref{equ:entropy_computation}.
This equation indicates that the entropy is a percentage ranged between 0.0\% and 100.0\%.

\begin{equation}
	entropy_{tw_{k}} = - \sum\limits_{i=1}^n \sum\limits_{j=1}^m \frac{p_{i,j}^{tw_{k}} \log_{2} p_{i,j}^{tw_{k}}}{\log_{2} (n \times m)} \times 100.0
\label{equ:entropy_computation}
\end{equation}

\begin{figure}
    \begin{minipage}[b]{0.53\textwidth}
        \centering \includegraphics[scale=0.2]{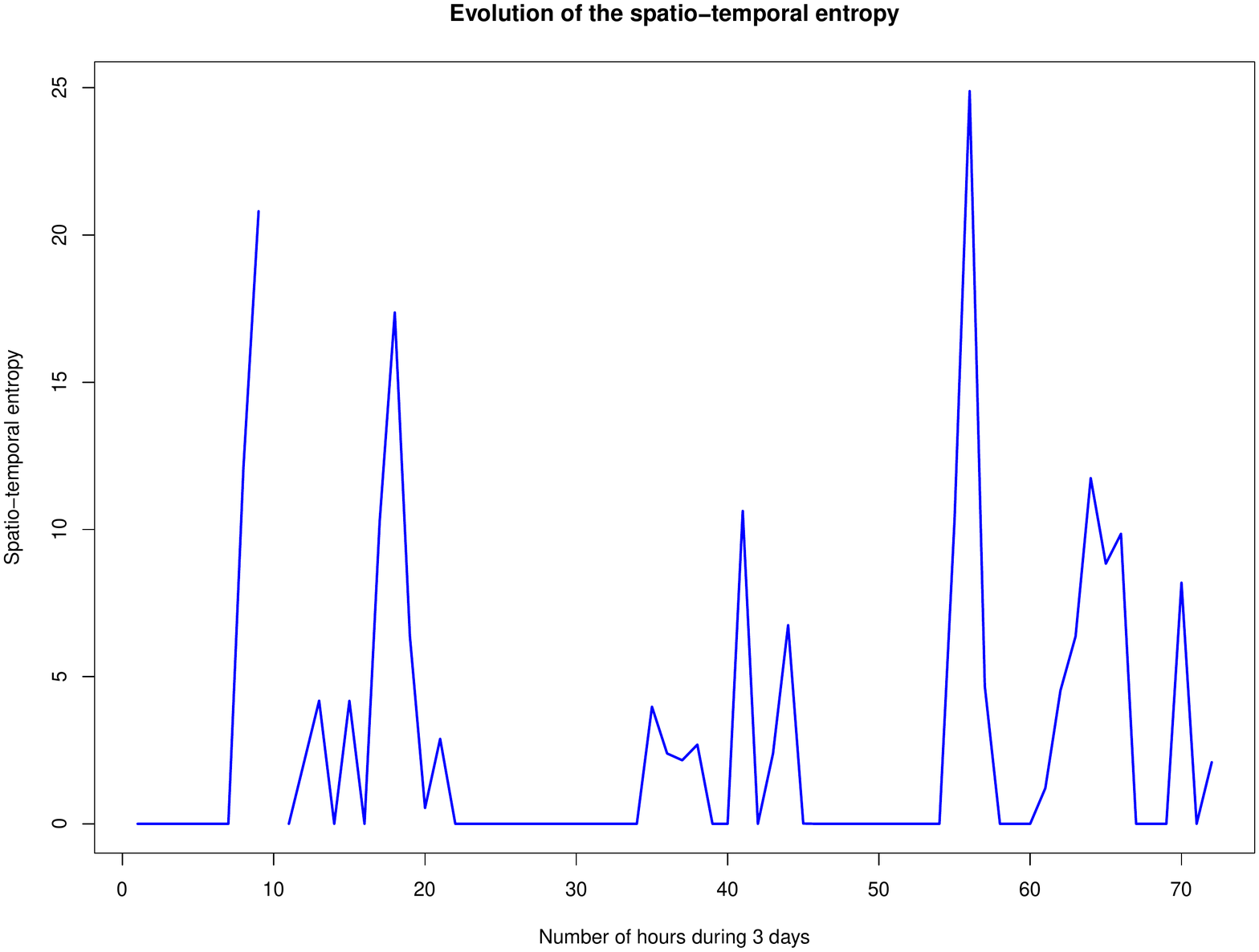}
    \end{minipage}
    \begin{minipage}[b]{0.53\linewidth}
        \centering \includegraphics[scale=0.2]{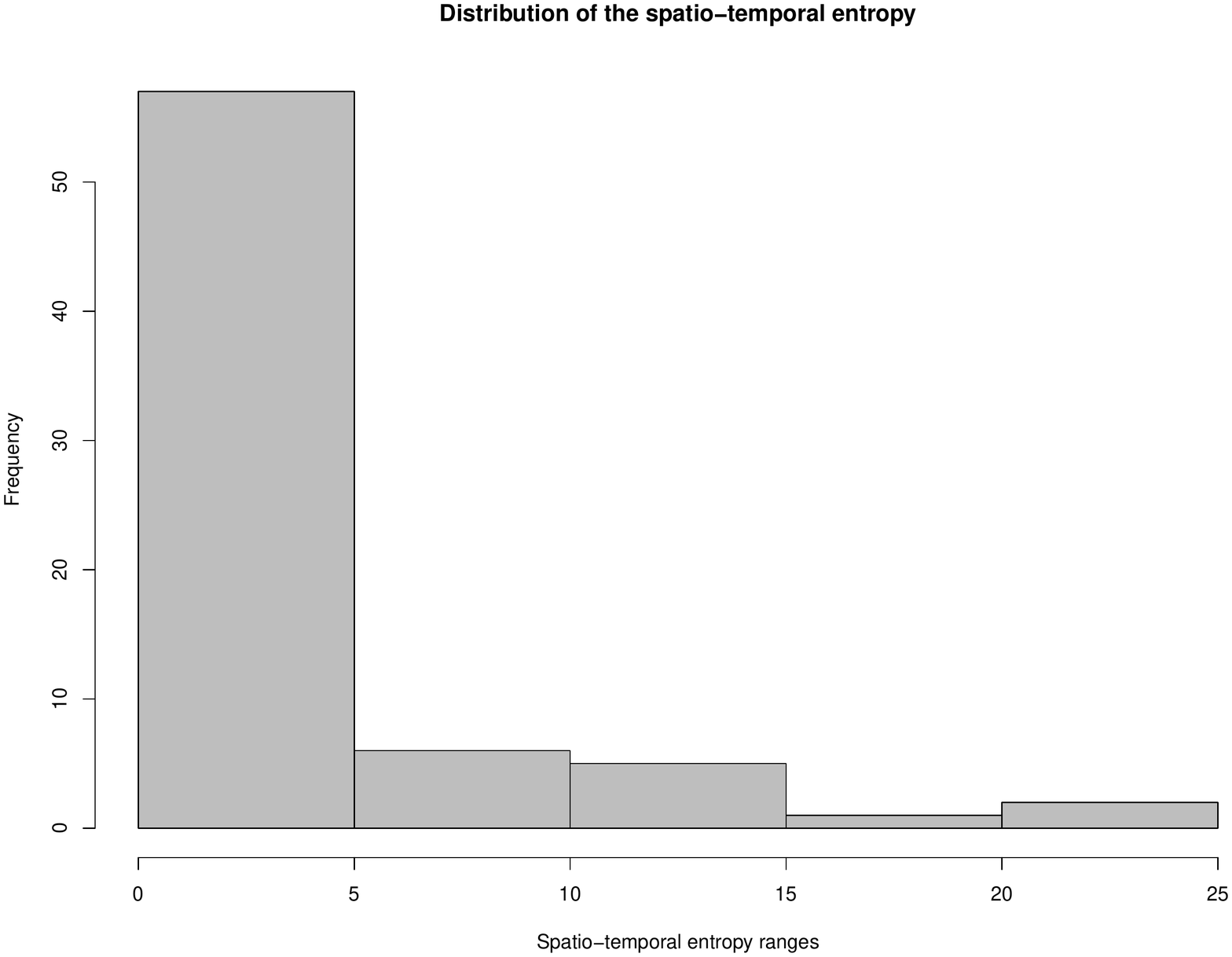}
    \end{minipage}
\caption{Spatio-entropy sequence detailed for one user during 3 days}
\label{fig:spatiotemporalentropy_evolution}
\end{figure}

At the end of this process, we must obtain a sequence of $T$ entropies that describe the movement rhythm represented by the spatio-temporal entropy sequence of the user.
Figure~\ref{fig:spatiotemporalentropy_evolution} (on the left) describes the evolution of the spatio-temporal entropy of a user of the data collection campaign \emph{Breadcrumbs} during three days.
We can observe stationary (when the entropy is equal to 0.0\%) and non-stationary periods (when the entropy is greater than 0.0\%).
In addition, Figure~\ref{fig:spatiotemporalentropy_evolution} (on the right) depicts the distribution of the spatio-temporal entropy during these three days for the same user.
We notice a high number of low entropy levels, this makes sense because a day of a typical human usually contains a large number of stationary periods per day, especially the users of the data collection campaign \emph{Breadcrumbs} who are mainly students (more information about \emph{Breadcrumbs} is available in Section~\ref{sec:dataset}).\\

\section{Mobility behavior analysis and prediction}
\label{sec:mobility_behavior_analysis_prediction}

We present in this section how it is possible to analyze mobility behavior of one user (a group of users) through the evolution of her (their) spatio-temporal entropy and additional variables that could help to better understand her (their) behavior.
The dataset used is called \emph{Breadcrumbs} and contains real mobility traces of users.
For this research work, we use three categories of variables that are detailed in Section~\ref{sec:variables_analyzed}: spatio-temporal variables, demographic variables and transportation modes preferences variables.

We first describe the dataset \emph{Breadcrumbs} we used and the variables we chose for this work.
Then, we demonstrate how it is possible to see the effects of these variables on a user's or users' movements expressed as an entropy rhythm with the use of Generalized Additive Models (GAMs).
Finally, we detail how we can predict this rhythm and what is the prediction accuracy we can obtain by using two types of models: GAMs and AutoRegressive Integrated Moving Average (ARIMA) models.

\subsection{Breadcrumbs dataset}\label{sec:dataset}

\emph{Breadcrumbs} dataset\footnote{Breadcrumbs Data Collection Campaign was a collaborative work of three research laboratories: Distributed Object Programming Lab (\url{http://doplab.unil.ch}), Information Security and Privacy Lab (\url{https://people.unil.ch/kevinhuguenin/}) and Business Information Systems and Architecture Lab (\url{https://wp.unil.ch/bisa/})} has been collected during a data collection campaign that started at the end of March 2018 and finished at the end of June 2018 in the Lake Geneva region, at the University campus of Lausanne more specifically (Switzerland).
We asked the participants (we called them users in the entire paper) to install an iOS application in order to record their location history during a period of three months.
The participants were mainly full-time students.
The reason why we decided to launch this data collection campaign was to obtain rich user datasets having a rich density in terms of location tracking: at least one location per hour in the best case.
In addition, we wanted to obtain a ground-truth of points of interest validated by the participants themselves.
For this specific research work, we selected~48 users having very rich location history with only~3 days without any location.
The duration of the selected user datasets is between~31 and~34 days.
Because the data collection campaign was in progress during the period of this analysis, we could only obtain this dataset duration.
To summarize, the added values of this dataset, compared to existing datasets, are that there are a higher location frequency tracking per day and also various additional variables that indicate demographic and mobility characteristics related to the participants, such as the transportation modes used, the age group, health information as well as a ground-truth of points of interest validated by the participants.
For sake of simplicity, the word \emph{user} is a synonym of \emph{participant} in this research paper.

\subsection{Selected variables}\label{sec:variables_analyzed}

As mentioned in the previous section, we also collected various additional data about demographic and mean of transport characteristics of the users of the \emph{Breadcrumbs} data collection campaign.
Moreover, we computed several spatio-temporal variables in addition to the spatio-temporal entropy for each time window analyzed for each user.
Below, you find the description of each selected variables for our research work and their possible values.\\

\noindent\textbf{Spatio-temporal variables (computed for each time window):}

\begin{itemize}
	\item tsnb: Number of the time window (over one week);
	\item maxdistance: Maximum distance travelled by the user during the time window;
	\item meanspeed: Mean speed of the user during the time window;
	\item maxspeed: Maximum speed of the user during the time window;
	\item campus: It the user is at the campus (No: 0 / Yes: 1);
	\item hourNb: Hour of the day (from 1 to 24);
	\item night: If the current is part of the night (night period: from 20 pm to 7 am);
	\item dayNb: Day of the week (from 1 to 7);
	\item prevdayNb: Previous day number of the current day of the week;
	\item nextdayNb: Next day number of the current day of the week;
	\item weekend: If the day is a weekend day (No: 0 / Yes: 1).
\end{itemize}

This first group of variables is generated for each computed time window.
As mentioned in Section~\ref{sec:from_locations_to_spatiotemporal_entropy}, if there is no location during a time window, the spatio-temporal entropy is equal to \emph{"NA"}.
Similarly, the spatio-temporal variables of a time window also have a value equal to \emph{"NA"} if there is no location is captured during this time window.\\

\noindent\textbf{Demographic variables (for each user analyzed):}

\begin{itemize}
	\item gender: Gender of the user (Male: 0 / Female: 1);
	\item age\_group: Age group of the user (Between 18 and 21 years old: 0 / Between 22 and 27 years old: 1 / Between 28 and 30 years old: 2 / More than 30 years old: 3);
	\item working\_profile: Working profile of the user (Studying full-time: 0 / Studying part-time :1 / Working part-time (less than 80\%) :2);
	\item job: If the user has a job during the week (No: 0 / Yes: 1);
	\item university: The university of the user (UniCampus1: 0 / UniCampus2: 1 / UniOutsideCampus1: 2 / UniOutsideCampus2: 3 / OtherUniversities: 4);
	\item section: Section of the user (Bachelor: 0 / Master: 1 / PhD: 2 / Other section: 3);
	\item living\_parent\_s\_home: If the user lives at her parents' home (No: 0 / Yes: 1);
	\item parent\_s\_home\_location: The location of the parents' home of the user (Other: 0 / Near the campus: 1);
	\item family\_status: Family status of the user (Single: 0 / Free union: 1 / Married: 2 / Divorced: 3 / Other: 4);
	\item sport\_exercises\_frequence: Sport activity frequency of the user (Less than 1 hour: 0 / Between 1 hour and 5 hours: 1 / More than 5 hours: 2);
	\item student\_association: If the user is the member of a student association (No: 0 / Yes: 1);
	\item smoking\_cigarettes: If the user smokes cigarettes (No: 0 / Yes: 1);
	\item seasonal\_allergies: If the user has seasonal allergies (No: 0 / Yes: 1);
	\item diet: The diet of the user (Diversified and not necessarily organic: 0 / Diversified and most of the time organic: 1 / Vegetarian: 2 / Vegan : 3 / Unspecified: 4).
\end{itemize}

\noindent\textbf{Transportation modes preferences variables (for each user analyzed):}

\begin{itemize}
	\item car\_week: If the user uses a car during the workweek (from Monday to Friday);
	\item car\_weekend: If the user uses a car during the weekend;
	\item public\_transportation\_week: If the user uses public transportation during the workweek;
	\item public\_transportation\_weekend: If the user uses public transportation during the weekend;
	\item bike\_week: If the user uses a bike during the workweek;
	\item bike\_weekend: If the user uses a bike during the weekend;
	\item taxi\_week: If the user uses taxi during the workweek;
	\item taxi\_weekend: If the user uses taxi during the weekend;
	\item walking\_week: If the user walks during the workweek;
	\item walking\_weekend: If the user walks during the weekend.
\end{itemize}

\subsection{Analyzing mobility behavior}

Below, we present the results obtained for the analysis of mobility behavior from the point of view of one single user and from the point of you of a group of users.
It is very important to indicate that the duration of each time window computed for each user is equal to 1 hour. 
The cells of the grid, used to compute the spatio-temporal entropy sequence, are generated by adding a latitude/longitude difference of 0.0025.
This means that the difference between two latitudes is approximately 278 meters and the difference between two longitudes is approximately 188 meters on an average for all cells of a common grid for the~48 users.
This very short duration is very flexible and enables to study seasonal movements: for example, every hour during one day, every day during one week, and for other seasonal scales depending on the entire dataset duration of a user.

\subsubsection{From one user point of view}

We observe firstly the impact of two seasonal variables on the mobility behavior of users of \emph{Breadcrumbs} dataset individually.
More specifically, our goal is to see how the spatio-temporal entropy is evolving at a scale of one day and at a scale of one week.
To do so, we build an individual GAM for each user, which is described in the following Equation~\ref{equ:seasonal_gam}, in which $y_{i}$ is the response variable, i.e., the spatio-temporal entropy, $\beta_{0}$ is the intercept and $\varepsilon_{i}$ the random error.
$s(hourNb, k=k{_{hour}})$ is a smooth function of the number of hour of the day and $s(dayNb, k=k{_{day}})$ is a smooth function of the number of day of the week, in which $k$ is a smoothing parameter representing the number of knots over the study period.
Figure~\ref{fig:spatiotemporalentropy_evolution} suggests that we use the Gamma family distribution for individual user spatio-temporal entropy.

\begin{equation}
	y_{i} = \beta_{0} + s(hourNb, k=k{_{hour}}) + s(dayNb, k=k{_{day}}) + \varepsilon_{i}
\label{equ:seasonal_gam}
\end{equation}
\vspace{0.2cm}

\begin{figure}
\center
\includegraphics[scale=0.34]{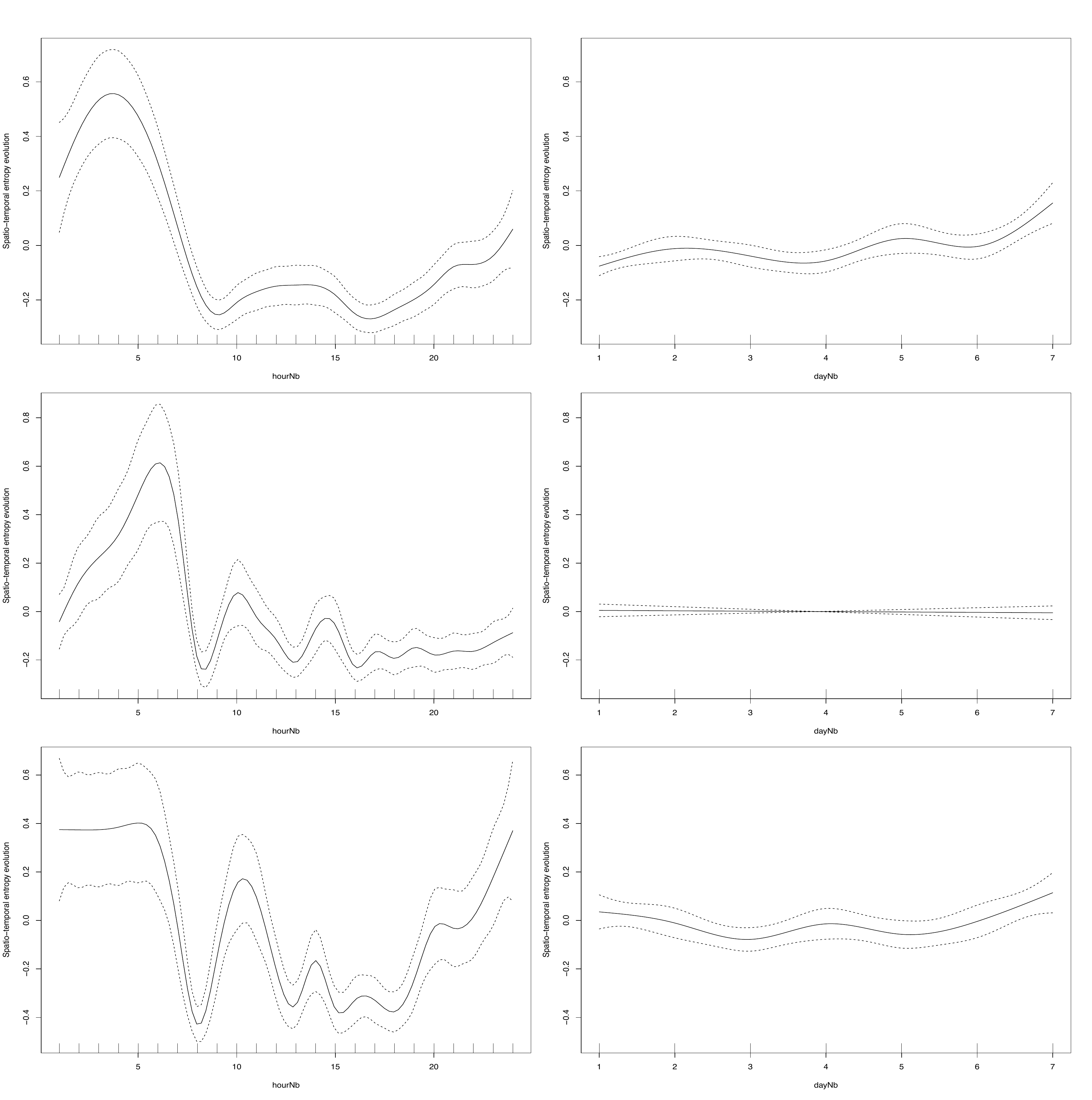}
\caption{Mobility behaviors of three different users according to seasonal variables}
\label{fig:seasonal_variables_effects_individual_users}
\end{figure}

\begin{figure}
\center
\includegraphics[scale=0.5]{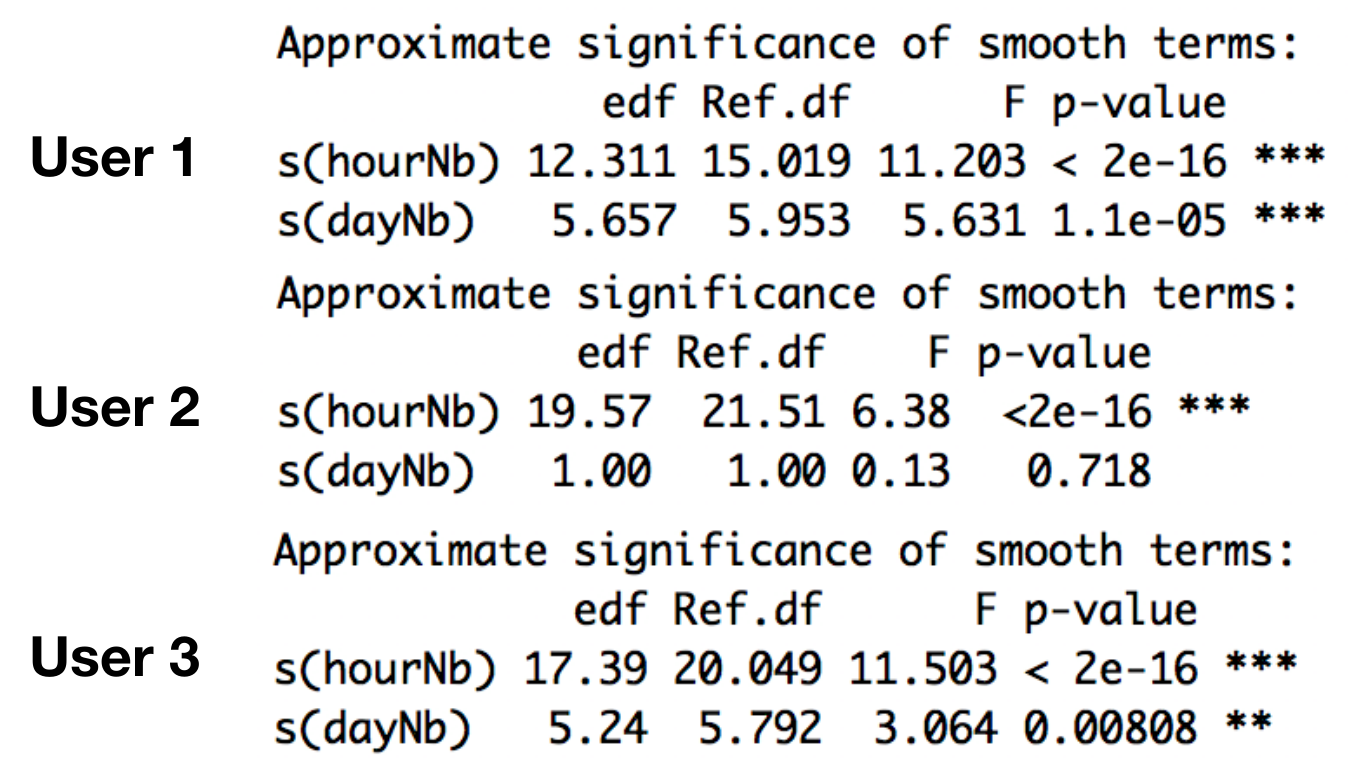}
\caption{Significance of the smooth terms for the three users}
\label{fig:smooth_terms_results}
\end{figure}

Figure~\ref{fig:seasonal_variables_effects_individual_users} describes an analysis of three different users of the \emph{Breadcrumbs} dataset (User 1 is at the top, User 2 is in the middle and User3 is at the bottom).
We can see on the left side, the evolution of the spatio-temporal entropy according to the hours of one day, and on the right side, the evolution of the spatio-temporal entropy according to the days of one week.
On the left side of the figure, if we look at the sub-figure of the User~2 in the middle, the entropy constantly increases and rapidly decreases after (maybe when the user reaches the campus) during the morning.
We observe that there are two peaks, in this same sub-figure, that could potentially highlight lecture breaks and the end of the day is relatively calm and constant.
On the right side of the figure, the three sub-figures clearly show different weekly mobility behavior, the week of User~2 in the middle is very regular compared to User~1 and User~3 for example.
In addition, Figure~\ref{fig:smooth_terms_results} highlights the significance of the smooth co-variables for each user by looking at the \emph{p-value}.
Without any surprise, the co-variable $hourNb$ is significant for all users, whereas $dayNb$ is only significant for User~1 and User~3.

\begin{figure}
\center
\includegraphics[scale=0.5]{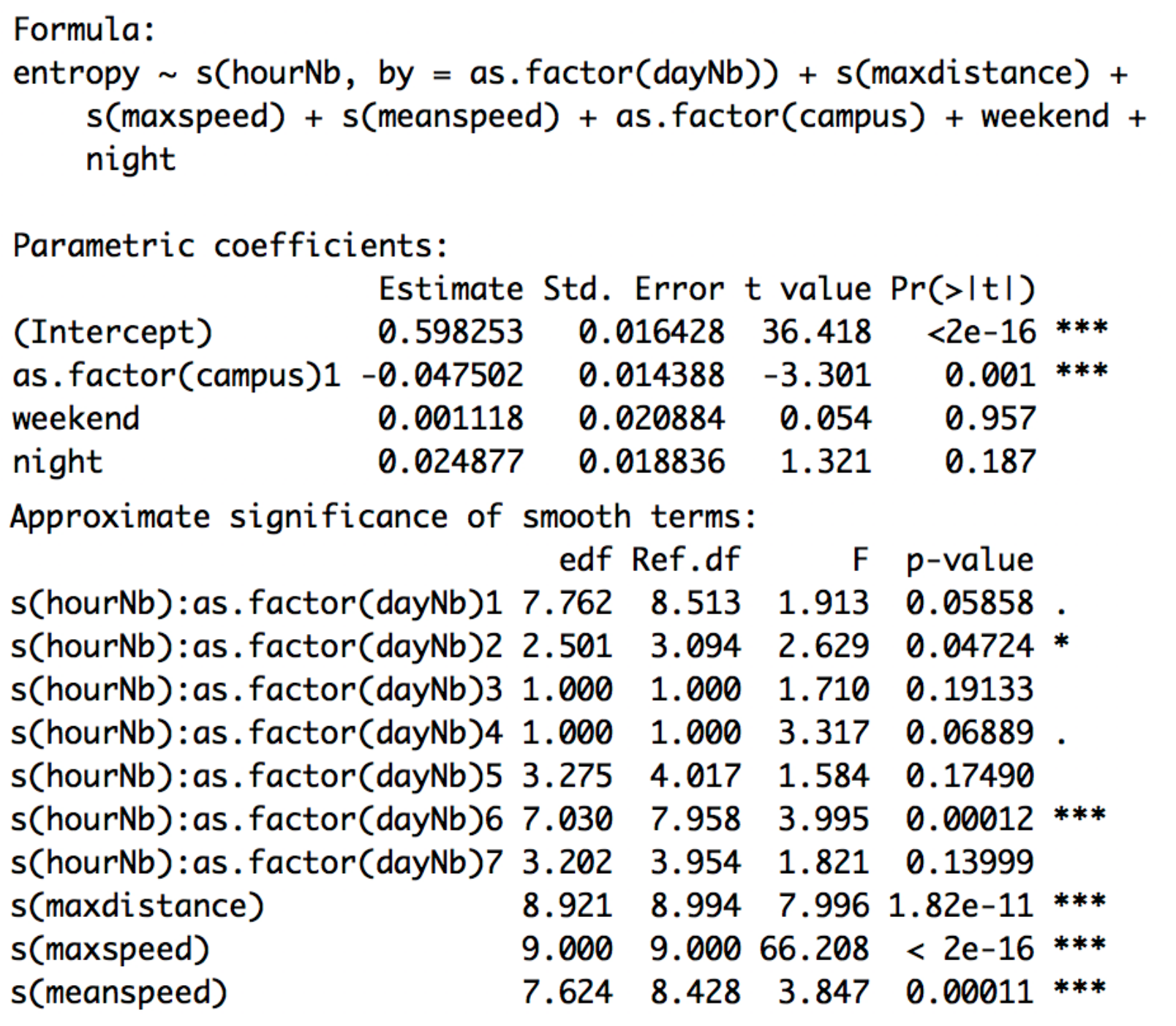}
\caption{Additional co-variables added in GAM formula for User 1}
\label{fig:more_complex_individual_gam}
\end{figure}

\begin{figure}
\center
\includegraphics[scale=0.34]{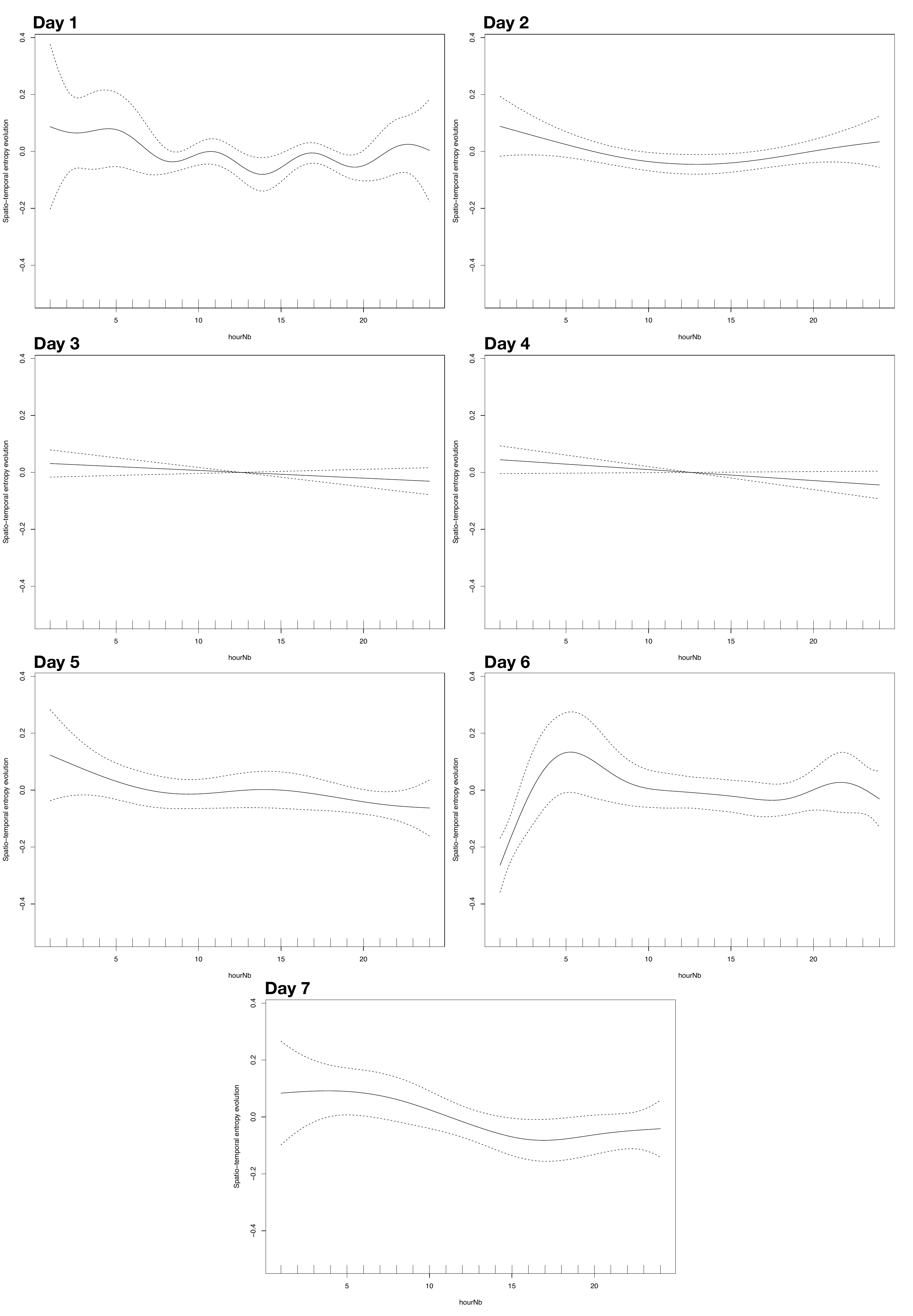}
\caption{Spatio-temporal evolution day after day for User 1}
\label{fig:user_1_entropy_evolution_day_by_day}
\end{figure}

Finally, Figure~\ref{fig:more_complex_individual_gam} shows the formula and the results of a more complex GAM applied on the data of User~1.
We can clearly note that, by seeing the $p-value$ of the co-variables, the spatio-temporal entropy decreases when the User~1 is at the campus.
The $p$-$values$ of the smooth terms are computed with a Wald test, while a test of Fisher (\emph{F}-test) is used to compute the $p$-$values$ of the parametric coefficients.
Figure~\ref{fig:user_1_entropy_evolution_day_by_day} depicts the evolution of the spatio-temporal entropy for every hour day after day.
From this figure, we can highlight different types of mobility per day, Day~3 and Day~4 are very regular compared to the other.
It is possible that User~1 stayed at home or at a fixed location during the days analyzed of her  dataset.
Interestingly, the weekend and the night does not necessarily shave an influence on the user's behavior.
Regarding the night, it is maybe due to the fact that the night period is tagged during a long duration (from 20 pm to 7 am) and that a user can move a lot during this period at home, especially if she is working late.
The Figure~\ref{fig:more_complex_individual_gam} also shows that the hours of Saturday (Day~6) have a specific influence on the evolution of the spatio-temporal entropy.
We also observe a high effect of the maximum distance, the maximum speed and the mean speed on the spatio-temporal entropy.

\subsubsection{From multiple users' point of view}

\begin{figure}
\center
\includegraphics[scale=0.5]{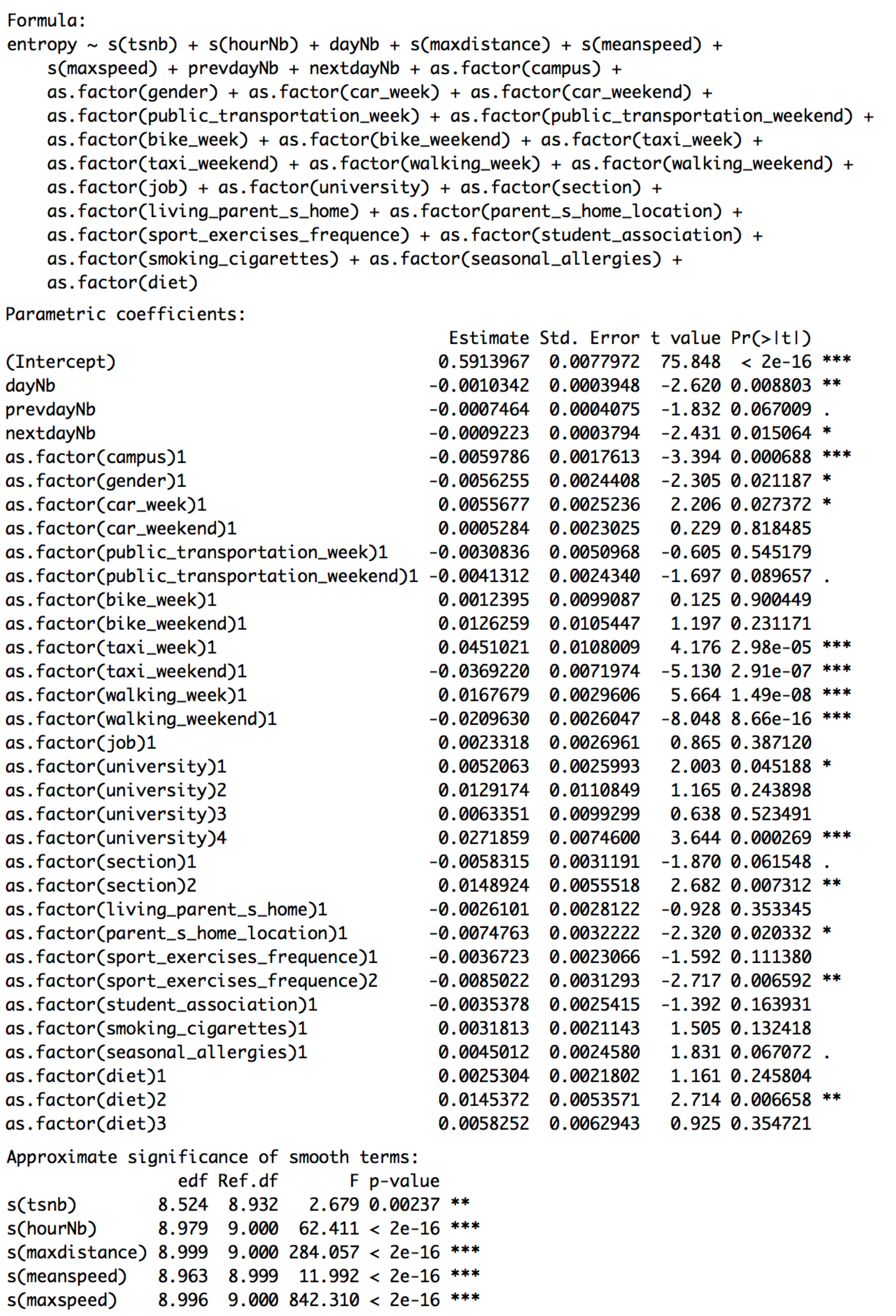}
\caption{Summary of the global GAM of 48 users}
\label{fig:summary_global_GAM_48users}
\end{figure}

This second analysis focuses on the mobility behavior of a group of users and, most specifically, the variables that can have an influence on their mobility.
For information, the distribution of the spatio-temporal entropy of the~48 users also follows a Gamma distribution, as for each individual user of the dataset.
Figure~\ref{fig:summary_global_GAM_48users} enables to see the influence of the spatio-temporal, demographic and means of transport variables of the evolution of the spatio-temporal entropy of the~48 users of the \emph{Breadcrumbs} dataset.
When we see different number for the different variables studied (as.factor(variablename)nb), this means that the variable is a factor with different levels.
From this figure, we are now able to highlight the co-variables that have an influence on the evolution of the spatio-temporal entropy for the~48 users.
Before talking about the results, we purposely removed some co-variables because some of their options were underrepresented.
Indeed, the~48 users were too homogeneous to analyze all demographic variables, such as age group, working profile and family status.
Hence, Figure~\ref{fig:summary_global_GAM_48users} depicts the results that are summarized below: 

\begin{itemize}
\item Being at the campus (campus 1), using a taxi during the weekend (taxi\_weekend 1), walking during the weekend (walking\_weekend 1) and doing a sport activity at a high frequency (sport\_exercices\_frequence 2) decrease the spatio-temporal entropy;
\item Using a taxi during the week (taxi\_week 1), walking during the week (walking\_week 1), studying in a university that is not located at the campus (university4), being in a master section (section 2) and having a specific diet (diet 2) increase the spatio-temporal entropy.
\end{itemize}

Interestingly, the results show that co-variable $dayNb$ has a negative influence on the evolution of the spatio-temporal entropy.
All these indicators give us good insights to better understand the way a specific population is moving.
These results also indicate that both demographic and means of transport co-variables can influence our mobility behavior.
Further work should be done to study the behavior of specific demographic groups with GAMs to explore them at a finer scale.


\subsection{Predicting mobility behavior}

In order to evaluate GAMs from a prediction perspective regarding the mobility of users, we decide to compare the prediction accuracy between GAMs and AutoRegressive Integrated Moving Average (ARIMA) models.

The GAMs predict a variable based on the value of the co-variables given, while the ARIMA models predict a variable based on its past values and its past variance according to three parameters~$p$, $d$ and~$q$.
Parameter $p$ is the number of autoregressive terms, parameter $d$ is the number of seasonal differences and parameter $q$ is the number of seasonal moving average terms.
For our research work, we entirely delegate the selection process of these three parameters by using a specific~$R$ function called $auto.arima()$ of the $forecast$ package.
An ARIMA model can be formally described as the following Equation~\ref{equ:arima_models}.

\begin{equation}
	y_{t'} = \mu + \phi_{1}y_{t-1} + \ldots + \phi_{p}y_{t-p} - \theta_{1}e_{t-1} - \ldots - \theta_{q}e_{t-q}
\label{equ:arima_models}
\end{equation}
\vspace{0.1cm}

In this equation, $\mu$ is a constant, $\phi_{1}y_{t-1} + \ldots + \phi_{p}y_{t-p}$ are the autoregressive terms and $\theta_{1}e_{t-1} - \ldots - \theta_{q}e_{t-q}$ are the moving average terms.
$y_{t'}$ is the integrated part with a certain difference noted $d$, such that:

\begin{itemize}
	\item if d = 0, $y_{t'} = y_{t}$ (no difference)
	\item if d = 1, $y_{t'} = y_{t} - y_{t-1}$ (first difference)
	\item if d = 2, $y_{t'} = (y_{t} - y_{t-1}) - (y_{t-1} - y_{t-2})$ (second difference)
	\item \ldots
\end{itemize}

\subsubsection{Evaluation process}

\begin{figure}
\center
\includegraphics[scale=0.5]{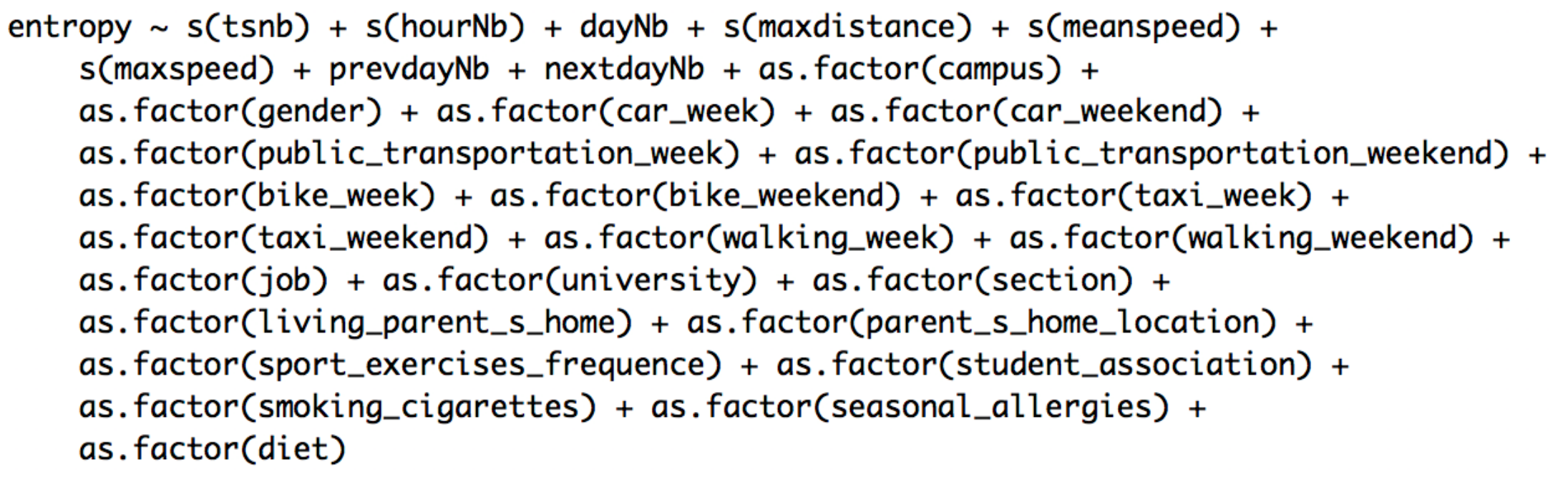}
\caption{Global GAM formula trained with 60\% of the dataset of the 48 users}
\label{fig:global_gam}
\end{figure}

For this evaluation, we decide to compare the prediction accuracy of a global GAM with all user data, individual GAMs for each user and individual ARIMA models for each user.
We firstly split the datasets of the users into two parts, the first part contains~60\% and the second part~40\% of the entire dataset of one user.
We train the global GAM (having the formula depicted in Figure~\ref{fig:global_gam}) with the aggregation of the first part of all user datasets and the individual GAMs (formula in  Figure~\label{fig:individual_gam}). and ARIMA models with each individual first part of the users.
Then, we try to predict the remaining~40\% part of each user dataset with the global GAM and the individual GAM and ARIMA model linked to the each user.
It is important to indicate that individual GAMs do not include demographic and means of transport variables of users, whereas the global GAM includes them.

For this comparison, we compute the Mean Absolute Error (MAE) and the Root Mean Squared Error (RMSE) of each predicted user set compared to the real values of the last~40\% remaining of the dataset of each user.
These two measures are complementary because the first metric highlights the absolute average magnitude of errors, whereas the second metric shows the standard deviation of these errors and gives more value to larger errors.

\subsubsection{Results}

Table~\ref{tab:global_gam_arima_mae_rmse_results} summarizes the results and shows that there is no a strong difference between the results obtained for the global GAM and the individual GAMs.
It is important to note that the results are expressed in the unit of the predicted variable, i.e., spatio-temporal entropy.
We can also note that the average of RMSE is lower for the global GAM compared to the individual GAMs.
This indicates that larger errors are maybe less numerous for the global GAM.
However, we can see that the average of the RMSE is slightly higher for the global GAM compared to that of the individual ARIMA models.
In addition, we can observe that the average of the MAE is almost similar for the global and individual GAMs, and slightly lower for the GAMs compared to that of the individual ARIMA models.

To conclude this prediction accuracy analysis, GAMs are more efficient to predict individual spatio-temporal entropy than ARIMA in terms of MAE.

\begin{figure}
\center
\includegraphics[scale=0.5]{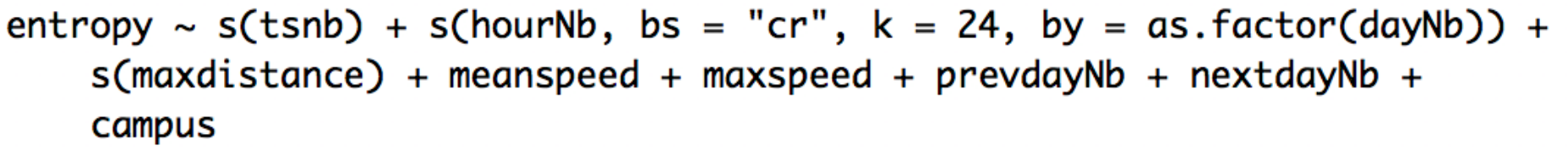}
\caption{Individual GAM formula trained with 60\% of the dataset of each user}
\label{fig:individual_gam}
\end{figure} 

\begin{table}
\scriptsize
\centering
\begin{tabular*}{0.909\textwidth}{|l|c|c|}
\hline
\textbf{Models} & \textbf{MAE (average for 48 users)} & \textbf{RMSE (average for 48 users)}\\
\hline
\hline
Global GAM & 2,74 & 5,45\\
\hline
Indiv. GAMs & 2,73 & 5,66\\
\hline
ARIMA models & 3,09 & 4,62\\
\hline
\end{tabular*}
\caption{Global GAM, individual GAMs and individual ARIMA models comparison: results}
\label{tab:global_gam_arima_mae_rmse_results}
\end{table}





\section{Related work}
\label{sec:related_work}

This section presents the related work close to our work according to its two main subjects: the analysis of mobility behavior and the use of entropy-based metrics in the literature.

\subsection{Mobility behavior analysis}

Most of the work realized in the literature about the analysis of mobility focus on mobility  patterns and use the significant locations visited by users.
Hence, it is possible to analyze the patterns extracted from the significant locations of users at different time scale and how they evolve over time.
In \cite{Gonzalez2008}, Gonzalez et al. analyze mobility patterns from the point of view of the significant locations visited by a user and observe the spatio-temporal regularity of user patterns.
In \cite{Zion2017}, Zion and Lerner also analyze the mobility of users by computing patterns with a preliminary analysis using the semantic of significant locations and a \emph{Latent Dirichlet Allocation} method personalized for temporal analysis.
In \cite{Sadilek2012}, Sadilek and Krumm perform a long-term analysis of patterns of users using Fourrier analysis and Principal Component Analysis.
They demonstrated the influence of temporal attributes, e.g., day of week, on user patterns.
Finally, in \cite{Kulkarni2016}, Kulkarni et al. analyze user mobility behavior in order to update mobility prediction models when major changes occur in the mobility of users.
In order to reach this goal, they analyze the frequency of movements and also the major changes that appear at the level of the significantly and frequently visited places by users.

\subsection{Entropy-based analysis}

The entropy is a well-known notion in the field of information theory, which can be applied to various domains, from physics to text analysis.
In the field of this work, i.e., human mobility analysis, the Shannon entropy is mainly used to evaluate the level of predictability of a user.
In \cite{Song2010}, Song et al. propose three entropy measures: the \emph{random entropy} capturing the degree of predictability of the user's locations, the \emph{temporal-uncorrelated entropy} highlighting the visit probability of a location by a user, and finally, the \emph{actual entropy} characterizing the level of order in the visit patterns of the user.
This paper demonstrates that users are highly predictable because their patterns seem to be very  regular.
The authors also observe that there are insignificant variations between different user groups, e.g., different genders, different age-groups.
The issue of this paper is that the dataset used was not very precise in terms of location tracking. 
More specifically, they use a hourly time window to compute the third entropy whereas a large number of hours was missing, 70\% to be precise for a typical user analyzed of the dataset.
And this low number of precise data maybe leads to an homogeneous vision of the mobility of the entire set of users.
In \cite{Qin2012}, Qin et al. also use the entropy to identify the level of predictability of a user.
They use a first entropy metric to measure the regularity level of a user over time slots.
In addition, they use a second entropy measure to evaluate the quality of the clusters because this can strongly have an influence on the prediction of the next significant location of a user.
The first measure of the entropy is mainly based on the significant location that dominates during a time slot in order to extract the regularity of the user over the entire time slots during one day at the end of the process.
In \cite{Lu2013}, Lu et al. compute the entropy in a three-step process as described in the first paper (see \cite{Song2010}) in order to highlight the level of predictability of users.
They discover that the population they study is highly predictable.
They also indicate the importance of weekly cycle in the mobility of their users.
As for the first paper mentioned in this section, they use a dataset in which the locations are cell phone towers.
The quality of the dataset used can also be a problem because a user can only have one location per day in the worst case.  
In \cite{Austin2014}, Austin et al. study the regularity and predictability of movements at home.
At this scale, they also find that the patterns of the users analyzed are significantly regular and open the work to the health domain.
In a very different domain, Sharma et al. study collaborative remote work amongst users in \cite{Sharma2016}.
They propose a computation of entropy according to a time proportion spent in different zones of a screen in order to analyze the effectiveness of collaborative remote work between two users.
Our computation of the spatio-temporal entropy is close to this last work and has been  personalized to our analysis.\\

\noindent To the best of our knowledge, the research works, about mobility and mentioned above, always use the most significant locations of users to study the mobility of the users. 
Our work opens a new entropy-based metric and a new complete methodology to study mobility in a privacy manner. 
Indeed, we avoid the step of the extraction of the significant locations of users in order to analyse the mobility of a user as a pure rhythm that can be influenced by several variables.
In addition, we use Generalized Additive Models (GAMs) to analyze the mobility rhythm of a user, and, to the best of our knowledge, these models were not applied on this field before.
We find their use in other scientific domain, such as environmental domain.
For example, in \cite{Pearce2011}, Pearce et al. analyse the impact of different local meteorology variables on air quality.

\section{Conclusion and future work}
\label{sec:conclusion}

This research work has two main contributions.
Firstly, it contributes to the description of a new methodology that enables to study user's movements in a privacy-aware manner without extracting sensitive places of users.
Secondly, we show that we are able to highlight the variables that can affect these movements using a sequence of spatio-temporal entropy that represents the user's movement as a rhythm.
A sequence of spatio-temporal entropy describes the level of mobility of a user as a time series, in which each spatio-temporal entropy is computed for a specific time window.
We also present the use of Generalized Additive Models (GAMs) for this mobility analysis by using a dataset we collected, called \emph{Breadcrumbs} during the data collection campaign of the same name.
This research work also highlights that a GAM can also be used to analyze the movements of an entire population, as we did it this work with the~48 students analyzed.
Compared to individual GAMs and ARIMA models, a global GAM provide the best prediction in terms of accuracy regarding the MAE.

This work opens three other interesting threads.
Firstly, we could use GAMs with longer user datasets, in terms of duration, in order to see if we observe lower MAE and RMSE and study their evolution over time for each user dataset.
This first analysis could also give a good understanding of the level of predictability of a user.
A second future work could focus on the prediction of the demographic data of a user based on the learning of the mobility trends of an aggregation of multiple datasets including the spatio-temporal entropy and other related variables.
Finally, the third work could be linked to the creation of individual personalized mobility prediction models for a pure location prediction goal.
These personalized models could be based on the seasonal mobility trends observed for each user in order to adjust their temporal scale.

\end{document}